\pdfoutput=1

\documentclass[11pt]{article}

\usepackage[]{acl}

\usepackage{times}
\usepackage{latexsym}
\usepackage{booktabs}
\usepackage[T1]{fontenc}

\usepackage[utf8]{inputenc}

\usepackage{microtype}
\usepackage{amsmath}
\usepackage{amssymb}
\usepackage{graphicx}
\usepackage{bm}
\usepackage{footmisc}
\usepackage{xspace}
\usepackage{cleveref}
\crefname{section}{\S}{\S\S}
\Crefname{section}{\S}{\S\S}
\crefname{table}{Tab.}{}
\crefname{figure}{Fig.}{}
\crefname{algorithm}{Algorithm}{}
\crefname{equation}{eq.}{}
\crefname{appendix}{Appendix}{}
\crefname{prop}{Proposition}{}
\crefformat{section}{\S#2#1#3} 

\newcommand{\sys}{CLP-Transfer\xspace}
\newcommand{\sysModelSUFFIX}{-CLP\xspace}

\title{Efficient Language Model Training through \\ Cross-Lingual and Progressive Transfer Learning}

\author{Malte Ostendorff \\
  DFKI GmbH \\
  Berlin, Germany \\
  \texttt{malte.ostendorff@dfki.de} \\\And
  Georg Rehm \\
  DFKI GmbH \\
  Berlin, Germany \\
  \texttt{georg.rehm@dfki.de} \\}

\begin{document}
\maketitle
\begin{abstract}

Most Transformer language models are primarily pretrained on English text, limiting their use for other languages.
As the model sizes grow, the performance gap between English and other languages with fewer compute and data resources increases even further.
Consequently, more resource-efficient training methods are needed to bridge the gap for languages with fewer resources available.
To address this problem, we introduce a cross-lingual and progressive transfer learning approach, called \sys, that transfers models from a source language, for which pretrained models are publicly available, like English, to a new target language.
As opposed to prior work, which focused on the cross-lingual transfer between two languages, we extend the transfer to the model size.
Given a pretrained model in a source language, we aim for a same-sized model in a target language.
Instead of training a model from scratch, we exploit a smaller model that is in the target language but requires much fewer resources.
Both small and source models are then used to initialize the token embeddings of the larger model based on the overlapping vocabulary of the source and target language.
All remaining weights are reused from the model in the source language.
This approach outperforms the sole cross-lingual transfer and can save up to 80\% of the training steps compared to the random initialization. 

\end{abstract}

\section{Introduction}

Large language models based on the Transformer architecture \citep{Vaswani2017} dominate today's NLP.
These models are typically pretrained on primarily English text \citep{Brown2020,Zhang2022a,Black2022}, except for a few multilingual models \citep{Scao2022,Lin2021a,Shliazhko2022}.
Given that multilingual models have been shown to perform suboptimal compared to monolingual ones \citep{Conneau2020,Nozza2020}, other languages than English benefit less from the recent progress in NLP.
As the model sizes grow, the performance gap between the models for English and other languages with fewer resources increases even further.
This gap is emphasized by \citet{Hoffmann2022}, as they show that model performance is not only bound by computing resources but mainly by data.
Consequently, more resource-efficient training methods are needed to bridge the gap for languages with fewer resources available.

Transfer learning is generally known to improve the training efficiency of various machine learning problems \citep{Tan2018ASO,Sun2018MetaTransferLF,Houlsby2019ParameterEfficientTL}. 
In the context of language models, efficient methods for task, language, or domain adaption have been proposed, e.g., adapters \citep{Chronopoulou2021,Pfeiffer2020}, bias term training \citep{Zaken2021}, and prompt tuning \citep{Guo2022ImprovingTS}. 
To obtain monolingual language models for low-resource languages, \citet{Minixhofer2021} and \citet{DeVries2021} have shown that available pretrained models, e.g., in English, can be recycled.
These cross-lingual transfer learning approaches reduce the training effort.
However, they only transfer across languages and neglect the sizes of the language models.
While training a large model may not be feasible in a low-resource setting, training a small or medium model is likely possible, as demonstrated by AraGPT2 \citep{antoun-etal-2021-aragpt2}, CamemBERT \citep{martin-etal-2020-camembert}, GPT-fr \citep{simoulin:hal-03265900}, GBERT \citep{chan-etal-2020-germans}, or Finnish BERT \citep{FinishBert}. 

This paper presents \sys, which is a cross-lingual and progressive transfer learning approach for language models.
As opposed to prior work, which focused on the cross-lingual transfer between two languages, we extend the transfer to the dimension of the model size.
Given a large and pretrained model in a source language, we aim for a same-sized model in a target language.
Instead of training the large model in the target language from scratch, we first train a smaller model that requires much fewer resources (or reuse a publicly available small model).
Both small and source models are then used to initialize the token embeddings of the large target model based on the overlapping vocabulary of the source and target language.
All remaining Transformer weights are reused from the large model in the source language.

We evaluate \sys for decoder-only language models based on GPT2 \citep{Radford2019} and BLOOM \citep{Scao2022}.
We use German as the target language.\footnote{German is not typically considered a low-resource language. However, at the time of writing, there are no monolingual German language models larger than 1B parameters publicly available. We will investigate other languages in the next iteration of this paper.}
The source models are either in English or multilingual.
Our experimental findings suggest that our approach outperforms the sole cross-lingual transfer and can save up to 80\% of the training steps compared to the random initialization.

\section{Related Work}

\paragraph{Cross-lingual Transfer.}

Exploiting pretrained models or data across languages is a common approach in NLP research \citep{zoph-etal-2016-transfer,lin-etal-2019-choosing,nguyen-chiang-2017-transfer}.
One challenge when transferring a model to a new language is preserving the capabilities of the source language.
\citet{artetxe-etal-2020-cross} proposed to replace the tokenizer and only train the token embeddings while freezing other Transformer layers of a multilingual BERT model.
Such a transfer approach produces monolingual models that can be independently fine-tuned to specific languages.
\citet{DeVries2021} followed a similar approach to transfer a GPT2 model to a new language.
Specifically, they transfer English GPT2 to Dutch and Italian by exclusively relearning the token embeddings and not the other model weights.
This forces the language model to learn token embeddings that are aligned between English and the target language.
However, freezing most parameters also limits the model's ability to learn about the new language.  
More recently,  \citet{Minixhofer2021} introduced the WECHSEL method that uses bilingual dictionaries to map the token embeddings from the source to the target language.
It reuses the Transformer weights from the source model and continues training them.
WECHSEL has been shown to outperform the transfer method from \citeauthor{DeVries2021}.

\paragraph{Progressive Transfer.}

Going from a small to a larger model is also known as progressive growing and was originally proposed to improve training stability.
\citet{Simonyan2014VeryDC} showed that starting from an efficient and small model and gradually increasing the model capacity yields more stable training.
The paradigm of progressive growth can also be used to accelerate model training which has been shown for various model architectures.
\citet{Karras2017} demonstrate this for GANs, \citet{Graves2016} for RNNs, and \citet{Gu2021} for BERT language models.
Furthermore, \citet{Gong2019} grow a BERT model in terms of its depth, i.e., they use trained weights of a shallow model to initialize a deeper model and achieve 25\% shorter training time.

\section{Methodology}

This section introduces the methodology that we follow for efficient language model training.

Our objective is to obtain a large language model $M_{t}^{\text{(large)}}$ with $p^{\text{(large)}}$ parameters for a target language $t$.
To increase the training efficiency, we omit the standard from-scratch training approach, i.e., randomly initializing the weights of  $M_{t}^{\text{(large)}}$.
Instead, our goal is to find a good initialization of the parameter weights of  $M_{t}^{\text{(large)}}$ such that training effort is reduced.
To achieve this, we exploit an already pretrained large language model $M_{s}^{\text{(large)}}$, also with $p^{\text{(large)}}$ parameters and the same model architecture but in a source language $s$, and a small pretrained language model $M_{t}^{\text{(small)}}$, with significantly fewer parameters $p^{\text{(small)}}<<p^{\text{(large)}}$ in the target language $t$.
The Transformer layer weights $\bm{W}_t$ from the large target model are initialized with the weights of $M_{s}^{\text{(large)}}$.
Similarily, token embedding weights $\bm{V}_t$ for that the tokens that exist in both the target and source language vocabulary are initialized with $\bm{V}_s$.
For the remaining token embeddings weights, a combination of $M_{s}^{\text{(large)}}$ and $M_{t}^{\text{(small)}}$ is used.
To get our approach to work, we rely on two assumptions about the vocabularies and token embedding spaces of the source and target language models.

\subsection{Assumptions}

Our approach makes the following assumptions:

\paragraph{Shared vocabulary.}
Our approach relies on the tokenizers of source and target languages sharing a substantial fraction of their vocabulary.
Given the tokenizer in the source and target language with their vocabularies $V_s$ and $V_s$, we assume that the number of tokens occurring in both vocabularies $V_s \cap V_s$ is significantly larger than zero, i.e., $|V_s \cap V_s|>>0$.\footnote{Another assumption is that tokenizers are identical across different model sizes as long as the language remains the same.}
Languages with the same script and from the same language family typically share more tokens.
For example, the overlap between German and English is higher compared to Arabic and English.
Notably, there will be always a certain overlap since Byte-Pair Encoding \citep{sennrich-etal-2016-neural} is the tokenization algorithm.
As shown in \cref{tab:sharedvocab}, our assumption holds for the source and target combinations tested in this paper.
While the English and German tokenizers share 24\% of their vocabulary, the multilingual BLOOM tokenizer also shares 5\% of the German vocabulary despite its much larger vocabulary size.

\begin{table}[h]

\centering
\footnotesize

\begin{tabular}{ll r}
\toprule
\textbf{Vocabulary $s$} & \textbf{Vocabulary $t$} & \textbf{$|V_s \cap V_t|$} \\ \midrule

English GPT2 &  German (ours)  &  24.04\% \\

Multilingual BLOOM  &  German (ours)  &  5.55\% \\

Multilingual XGLM &  German (ours)  &  2.62\% \\

English GPT2 &  Arabic GPT2  &  6.95\% \\

English GPT2  &  Finnish GPT2   &  13.71\% \\

Multilingual BLOOM & Arabic GPT2    & 6.52\% \\

Multilingual BLOOM & Finnish GPT2    & 3.54\% \\

\bottomrule
\end{tabular}

\caption{\label{tab:sharedvocab}Number of overlapping vocabulary tokens between different tokenizers, normalized by the source vocabulary size. The tokenizers are English GPT2  \citep{Radford2019}, Arabic GPT2 \citep{antoun-etal-2021-aragpt2}, Finnish GPT2\footnote{\url{https://huggingface.co/Finnish-NLP/gpt2-finnish}}, multilingual BLOOM  \citep{Scao2022}, multilingual XGLM \citep{Lin2021a}, and our German tokenizer.}

\end{table}

\paragraph{Token embeddings.}

A language model has the token embeddings $\bm{V} \in \mathbb{R}^{|V| \times h}$ that map each token $v$ in the vocabulary $V$ to its vector representation $\bm{v} \in \mathbb{R}^{h}$ with the hidden size of $h$.
For larger models, the hidden size $h$ of the token embedding is typically also larger compared to one of smaller models, i.e., $h^{\text{(large)}} > h^{\text{(small)}}$.
Despite varying in terms of $h$, we assume that relative positioning in the token embedding space remains comparable across model sizes.
The embeddings of a small model $\bm{V}^{\text{(small)}}$ would share spacial properties with the embeddings $\bm{V}^{\text{(large)}}$ of a large model.

\begin{figure}[ht]
\centering
\includegraphics[clip,width=0.85\linewidth]{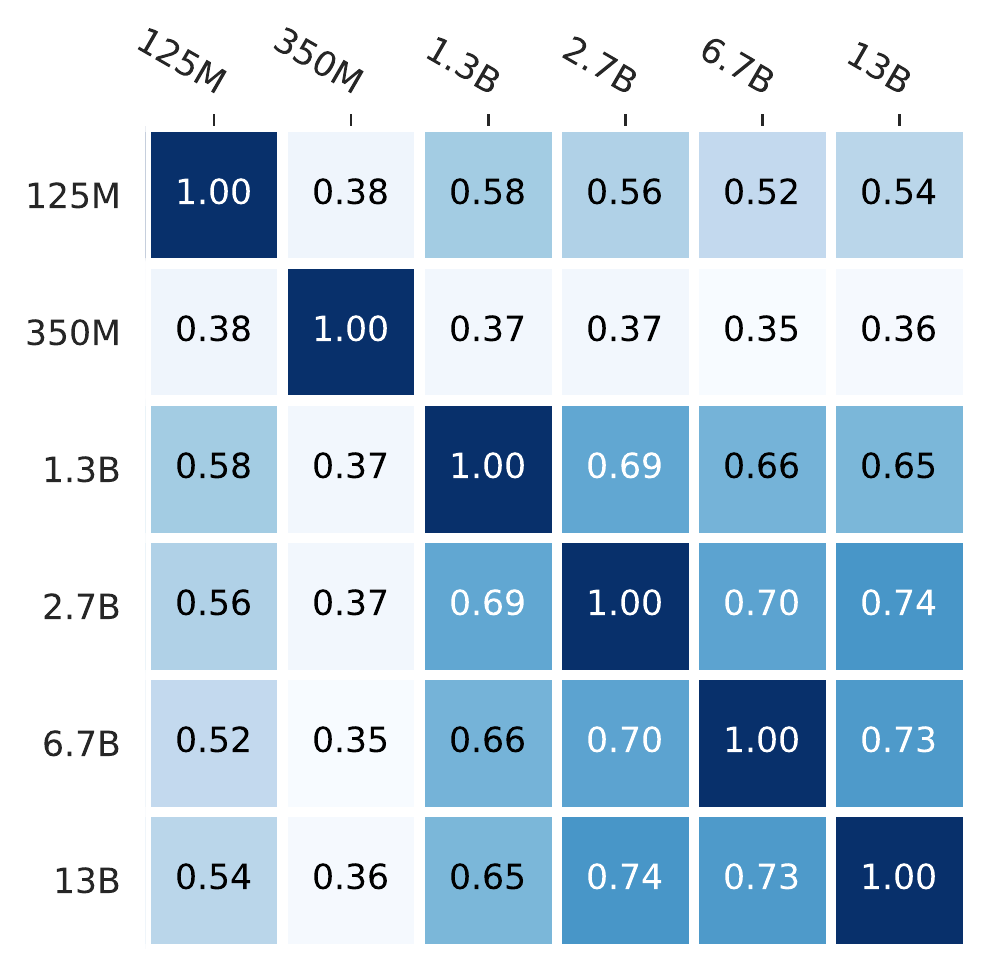}
\caption{\label{fig:token-embeddings}Similarity of token embeddings of different OPT model sizes measured as overlapping $k=10$ nearest neighbors for all tokens in the vocabulary.}
\end{figure}

To test this assumption, we compare token embeddings with different sizes from English OPT models \citep{Zhang2022a}.
Specifically, we retrieve the set of $k$-nearest neighbors $N_{v}$ with $k=10$ for each token $v$ and measure the overlapping neighbors for different model sizes, e.g., $N_{v}^{\text{(large)}} \cap N_{v}^{\text{(small)}}$.
This measure is normalized and computed for all available tokens.

As shown in \cref{fig:token-embeddings}, OPT token embeddings are comparable across model sizes.
The similarity between embedding spaces increases when the model size is comparable.
We find even between the smallest and the largest model (125M and 13B parameters) a 54\% overlap.
It is unclear why the 350M model has the lowest embedding similarity compared to all other models, independent of their size difference.

\subsection{Cross-lingual \& Progressive Transfer}

The weights of a language model in a language $i$ are comprised of token embeddings $\bm{V}_i$ and the Transformer weights $\bm{W}_i$.
We want to initialize $\bm{V}_t^{(large)}$ and $\bm{W}_t^{(large)}$ for our target language $t$ and the large model size.
The Transformer weights are simply initialized with the ones from the source language $s$:

\begin{equation}
\bm{W}_t^{(large)}=\bm{W}_s^{(large)}
\end{equation}

To initialize $\bm{V}_t^{(large)}$, we exploit $\bm{V}_s^{(large)}$ and $\bm{V}_t^{(small)}$, which are the token embeddings of a smaller model in the target language.
The embeddings of overlapping tokens that simultaneously exist in the source and target vocabulary  $v \in V_s \cap V_t$ are directly initialized with the source embeddings:

\begin{equation}
\bm{v_t}=\bm{v_s} \quad \text{if} \quad  v \in V_s \cap V_t
\end{equation}

When a token is not part of the overlapping vocabulary $v \notin V_s \cap V_t$, we initialize its embedding $\bm{v_t}$ as the weighted average over the embeddings $\hat{\bm{v}}$ of the overlapping token:

\begin{equation}
\begin{split}
\bm{v_t}^{\text{(large)}}= \sum\limits_{\hat{v} \in V_s \cap V_t} \frac{\hat{\bm{v}}_s^{\text{(large)}}}{\delta \left( v_t,\hat{v}_t \right) } \\
 \quad \text{if} \quad  v \notin V_s \cap V_t
\end{split}
\end{equation}

The weight function $\delta$ has the objective to transfer the spacial properties from the small model to the large model and is defined as the normalized cosine similarity of the small embeddings of overlapping $v$ and missing $\hat{v}$ tokens:

\begin{equation}
\delta(v,\hat{v}) = \frac{cos \left( \bm{v}_t^{(small)},\hat{\bm{v}}_t^{(small)} \right)}{ \sum\limits_{\substack{\hat{v}' \in V_s \cap V_t, \\ v' \in V_s \cup V_t}} cos \left( \bm{v}'^{(small)}_t,\hat{\bm{v}}'^{(small)}_t \right) }
\end{equation}

The intuition is those embeddings that are more similar in the $\bm{E}^{(small)}_t$ should contribute more to the construction of their corresponding token in the large model.
This approach allows us to recycle the pretrained weights of a source large model while preserving the spacial properties of the embedding space of the target language and simultaneously adjusting it to the vocabulary of our target language.

\section{Experiment Design}

In the experiments, we evaluate the \sys approach by transferring the English GPT2 \citep{Radford2019} and multilingual BLOOM \citep{Scao2022} to a monolingual German language model.
Both model types are evaluated at different scales.
More specifically, we grow the GPT2 model from 117M to 1.5B parameters and the BLOOM model from 1.5B to 6.4B parameters.

\subsection{Models}

\paragraph{Model Architectures.}

Both models (GPT2 and BLOOM) are decoder-only language models based on the Transformer architecture \citep{Vaswani2017} and are trained with the causal language modeling objective.
GPT2 uses learned positional embeddings, whereas BLOOM uses ALiBi \citep{Press2022}.
Another difference is that BLOOM applies normalization on the token embedding layer to improve training stability.

In our experiments with GPT2, we aim for a monolingual German model with the size of GPT2-XL with 1.5B parameters.
The source model is the English GPT2-XL as provided by \citet{Radford2019}\footnote{\label{fn:gpt2-xl}\url{https://hf.co/gpt2-xl}}
with 48 layers, 25 attention heads, and a hidden size of 1600.
As small German model, we use a GPT2-base model with 117M parameters trained with WECHSEL \citep{Minixhofer2021}.\footnote{\label{fn:gpt2-wechsel-german}\url{https://hf.co/malteos/gpt2-wechsel-german-ds-meg}}
The small German model has 12 layers, 12 attention heads, and a hidden size of 768.

To test our transfer method for another model type and size, we also conduct experiments with BLOOM.
For this experiment, our objective is the training of a German model based on BLOOM with 7.1B parameters as the source model.\footnote{\label{fn:bloom-7b1}\url{https://hf.co/bigscience/bloom-7b1}}.
The BLOOM 7.1B model has 30 layers, 32 attention heads, and a hidden size of 4096.
Our German BLOOM target model uses a different tokenizer with a smaller vocabulary size (see below). 
Therefore, its token embedding layer contains fewer parameters than the multilingual BLOOM model. 
As a result, the target model has only 6.4B parameters.
The small German model is a BLOOM model with 1.5B parameters trained with our method (24 layers, 16 attention heads, and a hidden size of 2048).\footnote{\label{fn:bloom-1b7-twc-german}\url{https://hf.co/malteos/bloom-1b5-clp-german}}

\paragraph{Tokenizers.}

All used tokenizers are based on Byte-Pair Encoding \citep{sennrich-etal-2016-neural}.
The vocabulary size of English GPT2 is 50,257 tokens.
BLOOM is multilingual and covers a diverse set of 46 natural languages and 13 programming languages.
Therefore, BLOOM's vocabulary size is 250,880 tokens, five times larger than the one from English GPT2.
For our German tokenizer, we opt for the same vocabulary sizes as the English one (50,257 tokens) and train it on the German subset of OSCAR v2019 \citep{OrtizSuarezSagotRomary2019}.

\subsection{Datasets}

The GPT2 and BLOOM models are trained with two different datasets.

\paragraph{GPT2 Training.}

The German GPT2\sysModelSUFFIX training relies exclusively on Web-crawled data from the German subset of OSCAR v2019 \citep{OrtizSuarezSagotRomary2019}.\footnote{\url{https://hf.co/datasets/oscar} (subset: \texttt{unshuffled\_deduplicated\_de})}
We follow the methodology from \citet{Minixhofer2021} to construct a separate training and validation dataset.
Specifically, we used the first 4 GB of OSCAR as the training dataset, then the next 0.4GB as the validation dataset. 
The GPT2 training dataset comprises approximately 30.8B tokens.

\paragraph{BLOOM Training.}

To train the German BLOOM\sysModelSUFFIX 6.4B model, we construct another dataset.
We use again Web-crawled content from the German subset OSCAR but the more recent version of v22.01 (excluding content tagged as \textit{header}, \textit{footer}, \textit{noisy}, or \textit{adult}) and from the GC4 Corpus\footnote{\url{https://german-nlp-group.github.io/projects/gc4-corpus.html}} (including only the \textit{head} and \textit{middle} parts).
As both data sources originate from CommonCrawl and potentially have duplicated content, we deduplicate the Web-crawled content using the approach from \citet{Lee2022}.
We complement the Web-crawled content with German court decisions from Open Legal Data \citep{Ostendorff2020f}. 
The BLOOM training dataset comprises approximately 50.4B tokens.

\paragraph{Evaluation Datasets.}

We evaluate the models for their language modeling ability using the OSCAR validation set from the GPT2 training\footnote{\label{fn:oscar-validation}\url{https://hf.co/datasets/malteos/wechsel_de}}, and for zero-shot learning on German downstream tasks.
The tasks are sentiment analysis from GermEval 2017 \citep{wojatzki2017germeval}, hate speech classification from GermEval 2018 \citep{Wiegand2018OverviewOT}, news topic classification from GNAD10 \citep{Schabus2017}, paraphrase identification from PAWSX \citep{yang-etal-2019-paws}, natural language inference from XNLI \citep{conneau2018xnli}, and stance detection from X-Stance \citep{vamvas2020xstance}.
All evaluation tasks are implemented using the \texttt{lm-evaluation-harness} framework \citep{eval-harness}.\footnote{\url{https://github.com/OpenGPTX/lm-evaluation-harness}}

\subsection{Baselines}

We compare against the following baselines:

\paragraph{From-Scratch Training.} The language model is trained from scratch in the target language with randomly initialized weights.

The from-scratch baseline for the BLOOM experiments (BLOOM 6.7B) was trained with minor changes to the transferred BLOOM\sysModelSUFFIX 6.4B model.  
The baseline BLOOM 6.7B follows the model size proposed by \citet{Brown2020}. 
It has 32 layers instead of 30 layers and was not trained on GC4 and Open Legal Data but on other German datasets.

\paragraph{WECHSEL.} The WECHSEL method as introduced by \citet{Minixhofer2021} applies cross-lingual transfer to monolingual language models.
WECHSEL uses bilingual dictionaries to map the token embeddings from a source language to a target language and reuses the Transformer weights from the source model.
WECHSEL has been shown to outperform the transfer method from \citet{DeVries2021}.

\paragraph{Multilingual Models.} Lastly, we compare the monolingual German models against multilingual models trained on German data.
We evaluate XGLM \citep{Lin2021a} ranging from 564M to 7.5B parameters and mGPT \citep{Shliazhko2022} with 1.3B parameters.
XGLM was trained on approx. 5.4\% German data and mGPT on 8.2\% German data.

\section{Results}

We show the training results of two monolingual German language models, i.e., GPT2 1.5B and BLOOM 6.2B.
The models are evaluated regarding their training efficiency and downstream task performance.

\subsection{Transfering GPT2}

\begin{figure}[ht]
\centering
\includegraphics[clip,width=1.\linewidth]{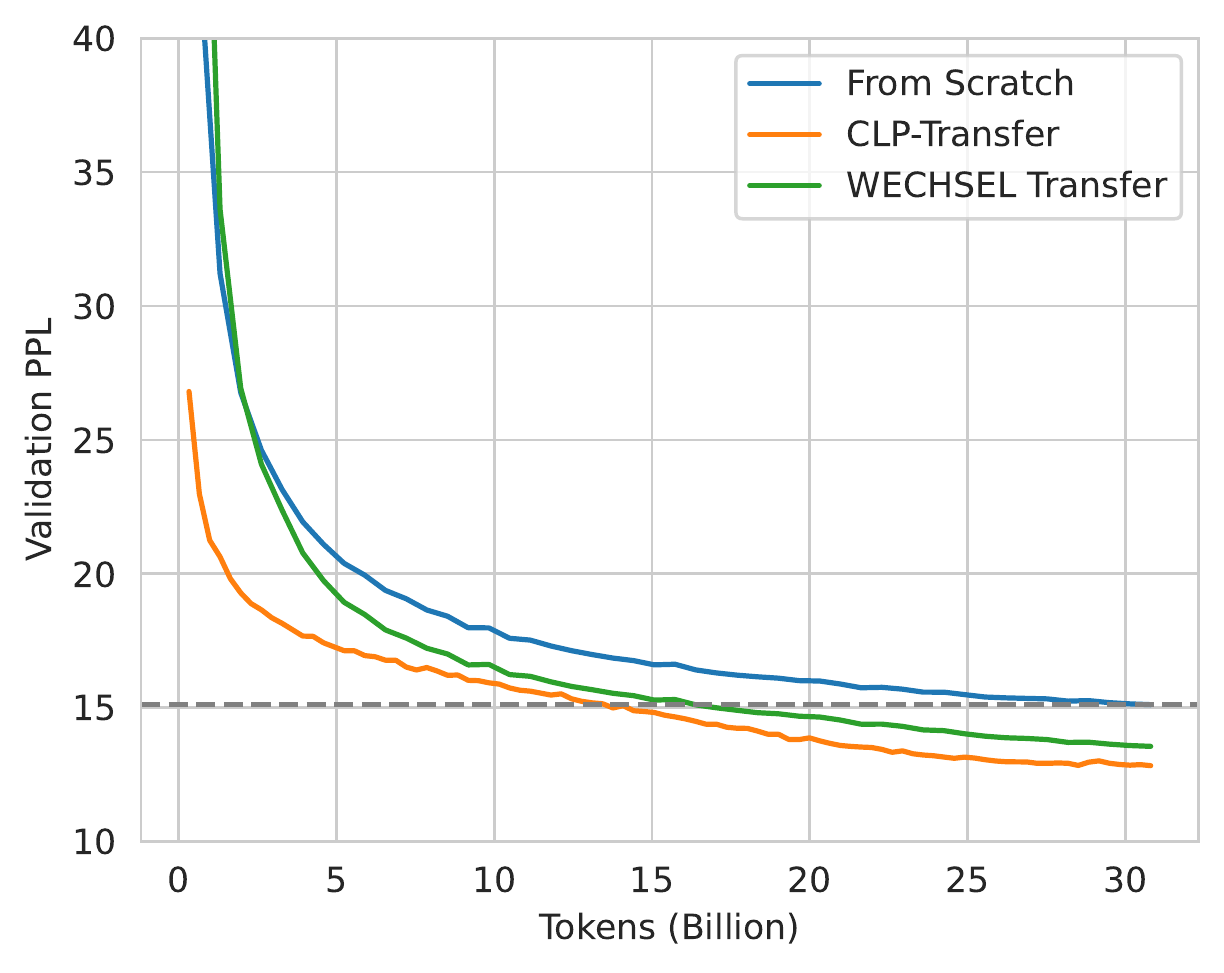}
\caption{\label{fig:gpt2-ppl}GPT2-XL German (1.5B parameters). Validation perplexity w.r.t. the number of consumed tokens comparing from-scratch training (random initialization), WECHSEL (cross-lingual transfer), and our \sys approach. \sys achieves the same PPL as from-scratch training but already after 50\% of tokens (dashed line).}
\end{figure}

The goal of our first experiment is the training of a German GPT2-XL model with 1.5B parameters.
\sys is compared against from-scratch training and the cross-lingual transfer method from WECHSEL.
\cref{fig:gpt2-ppl} shows the validation perplexity (PPL) of each method in relation to the training progress measured in consumed tokens.

We find that \sys outperforms the baselines. 
The validation PPL of \sys is constantly the lowest of all three methods. 
At the end of the training (after 30.8B tokens),  \sys yields a 12.8 PPL, followed by WECHSEL with 13.5 PPL.
The worst result has the from-scratch training with 15.1 PPL.
\sys achieve the same PPL as from-scratch training but already have 50\% of the consumed tokens.
During the first phase of the training (0-5B tokens), the improvements of \sys are most significant.
These results demonstrate that our transfer learning approach is superior to from-scratch training even at the end of the training or can achieve the same results more efficiently.
Moreover, the additional use of a small model in the target language yields further efficiency gains compared to the sole cross-lingual transfer done by WECHSEL.

\subsection{Transfering BLOOM}

\begin{figure}[ht]
\centering
\includegraphics[clip,width=1.\linewidth]{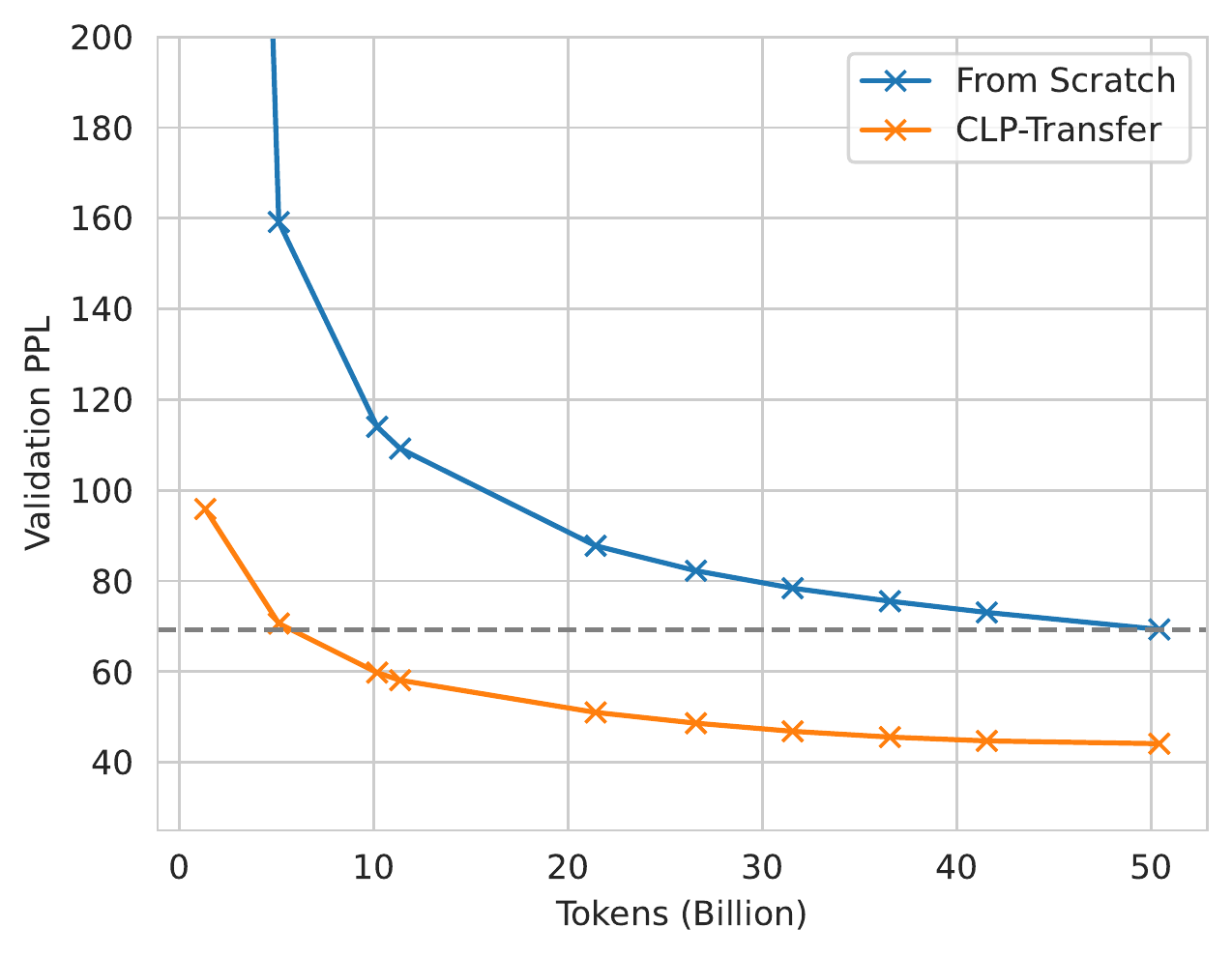}
\caption{\label{fig:german6b-ppl}BLOOM-6B-German. Validation perplexity w.r.t. the number of tokens comparing from-scratch training (random initialization) and our \sys approach. \sys achieves the same PPL as from-scratch training but after 20\% of tokens (dashed line).}
\end{figure}

The second experiment applies \sys on a multilingual BLOOM model\footref{fn:bloom-7b1} to train a monolingual German model with 6.4B parameters.
In this experiment, we compare only against from-scratch training. 
We discarded WECHSEL from this experiment due to the results from the first experiment and due to WECHSEL being made for the transfer of monolingual models and not multilingual ones. 

As shown in \cref{fig:german6b-ppl}, \sys again outperforms the from-scratch training.
After complete training on 50.4B tokens, \sys yields a 44.1 PPL, whereas from-scratch training is significantly worse with 69.3 PPL.
20\% of training tokens are sufficient for \sys to be on par with from-scratch training.
This suggests that the efficiency gains from \sys are even more prevalent at 6B compared to 1.5B parameters.
We attribute this outcome to the training data containing too few tokens for 6B models.
The validation PPL is still decreasing at the end of the training suggesting that the model is not fully trained yet.
According to \citet{Hoffmann2022}, a compute-optimal language model at the 6B scale would require approx. 142B tokens which our BLOOM model training did not consume.
The GPT2 training is much closer to being compute-optimal (\citeauthor{Hoffmann2022} suggest 33B tokens for a 1.5B model).

\begin{table*}[!ht]
\caption{\label{tab:zershot}Evaluation results of German downstream tasks in a zero-shot setting. The average score excludes the OSCAR validation perplexity (PPL). Smaller models are on par or worse than the random baseline. Our transfer model BLOOM\sysModelSUFFIX 6.4B achieves the best results on average.}
\centering
\footnotesize
\label{tab:my-table}
\setlength{\tabcolsep}{3pt}
\begin{tabular}{lr cccc ccc}
\toprule
\textbf{Task} $\rightarrow$           & \textbf{Oscar} & \textbf{GEval17} & \textbf{GEval18} & \textbf{GNAD10} & \textbf{PAWSX} & \textbf{XNLI} & \textbf{XStance} & \textbf{Avg.} \\

\textbf{Model} $\downarrow$  / \textbf{Metric} $\rightarrow$  & PPL ($\downarrow$)   & F1 ($\uparrow$)        & F1 ($\uparrow$)         & F1 ($\uparrow$)     & F1 ($\uparrow$)    & Acc. ($\uparrow$) & F1 ($\uparrow$)      &    ($\uparrow$)   \\ \midrule

Random                     &          - &          0.33 &                 0.50 &    0.11 &      0.50 &     0.33 &         0.50 &  0.38 \\

\addlinespace[1ex]
\multicolumn{9}{l}{\textit{Multilingual models:}} \\

mGPT 1.3B                  &       2274.80 &          0.36 &                 0.51 &    0.08 &      0.49 &     0.37 &         0.49 &  0.38 \\
XGLM 564M                  &        179.59 &          0.05 &                 0.40 &    0.05 &      0.46 &     0.44 &         0.50 &  0.32 \\
XGLM 1.7B                  &        105.10 &          0.04 &                 0.35 &    \textbf{0.19} &      \textbf{0.58} &     0.45 &         0.40 &  0.34 \\
XGLM 7.5B                  &         66.74 &          0.51 &                 0.51 &    0.06 &      0.50 &     0.39 &         0.41 &  0.40 \\

\addlinespace[1ex]
\multicolumn{9}{l}{\textit{Monolingual German models:}} \\

GPT2-WECHSEL 117M          &        594.40 &          0.04 &                 0.51 &    0.18 &      0.49 &     0.40 &         \textbf{0.51} &  0.35 \\
GPT2-XL-WECHSEL 1.5B       &        157.95 &          0.05 &                 \textbf{0.55} &    0.10 &      0.41 &     \textbf{0.49} &         0.34 &  0.32 \\
GPT2-XL\sysModelSUFFIX 1.5B            &         46.33 &          0.05 &                 0.02 &    0.07 &      0.46 &     \textbf{0.49} &         0.34 &  0.24 \\
GPT2-XL 1.5B from scratch              &        187.71 &          0.04 &                 0.51 &    0.15 &      0.52 &     0.47 &         0.34 &  0.34 \\

\addlinespace[1ex]
BLOOM\sysModelSUFFIX 1.5B              &         49.80 &          0.04 &                 0.14 &    0.11 &      0.44 &     0.48 &         0.38 &  0.26 \\
BLOOM\sysModelSUFFIX 6.4B (50B tokens) &         \textbf{44.09} &          \textbf{0.56} &                 0.51 &    0.13 &      0.52 &     0.43 &         0.44 &  \textbf{0.43} \\
BLOOM 6.7B from scratch (50B tokens)    &         69.32 &          0.51 &                 0.52 &    0.13 &      0.41 &     0.38 &         0.42 &  0.39 \\
BLOOM 6.7B from scratch (72B tokens)    &         64.03 &          \textbf{0.56} &                 0.51 &    0.09 &      0.40 &     0.37 &         0.49 &  0.40 \\
\bottomrule
\end{tabular}

\end{table*}

\subsection{Downstream Tasks}

Even though we trained the models exclusively with a causal language modeling objective, we want them to perform well on other downstream tasks, as shown by \citet{Brown2020}.
Hence, we compare the models and additional baselines on six German benchmarks in a zero-shot setting.
Given that the from-scratch trained BLOOM 6.7B model (50B tokens) is presumable under-trained, we add an additional variation that was trained on 22B more tokens, i.e., BLOOM 6.7B (72B tokens).
The evaluated tasks are sentiment analysis (GermEval 2017), hate speech classification (GermEval 2018), news topic classification (GNAD10), paraphrase identification (PAWSX), natural language inference (XNLI), and stance detection (X-Stance).
\cref{tab:zershot} reports the validation PPL on German OSCAR\footref{fn:oscar-validation}, the results for the individual tasks, and the average over the tasks.

The zero-shot performance of all models is disappointing.
Most models achieve results on par or worse than the random baseline.
Only the largest models (more than 6B parameters) are better than the random baseline on average.
The BLOOM\sysModelSUFFIX 6.4B model has the best average score of 0.43, followed by the from-scratch trained BLOOM 6.7B (72B tokens) and XGLM 7.5B.

We hypothesize that this outcome is due to the model size and token count being still too small. 
Studies from \citet{Black2022} or \citet{Shliazhko2022} report similar near-random results for models with comparable sizes.
Another reason might be the poorly translated test datasets that produce less meaningful results. 
For instance, PAWSX contains a large fraction of machine-translated samples.
To improve the downstream task performance, promising approaches are prompt engineering, i.e., tailoring the prompts more to the German language, and multi-task fine-tuning, as demonstrated by BLOOMZ \citep{Muennighoff2022a} or FLAN \citep{Wei2022}.

\section{Conclusion}

This paper introduces \sys, which is a cross-lingual and progressive transfer learning approach for the efficient training of large language models.
Our experiments demonstrate that monolingual German language models initialized with \sys reduce the training effort.
The \sys models achieve better results when trained on the same number of tokens than from-scratch training or WECHSEL transfer.
To obtain the same perplexity as from-scratch training, \sys needs only 50\% (GPT-2) or even 20\% (BLOOM) of the original token count.
This corresponds to a 50\% or 80\% reduction in training effort.
Such a reduction lowers the barriers to the training of large language models in low-resource settings.

The training efficiency is achieved by exploiting publicly available models, i.e., English or multilingual large models and small models in the target language.
\sys relies on the assumptions that vocabularies of source and target language have significant overlap and that small and large model have similar token embedding spaces.

We make the pretrained model checkpoints and our source code publicly available on HuggingFace\footnote{\url{https://huggingface.co/malteos/bloom-6b4-clp-german}} and GitHub\footnote{\url{https://github.com/malteos/clp-transfer}}.
Furthermore, we provide a Web-based demo in which the German BLOOM-CLP model with 6.4B parameters can be prompted.\footnote{\url{https://ostendorff.org/clp}}

\section*{Acknowledgements}

The work presented in this paper has received funding from the German Federal Ministry for Economic Affairs and Climate Action (BMWK) through the project \href{https://opengpt-x.de/}{OpenGPT-X} (project no. 68GX21007D).

The authors gratefully acknowledge the \href{https://www.gauss-centre.eu}{Gauss Centre for Supercomputing e.V.} and the GWK support for funding this project by providing computing time through the John von Neumann Institute for Computing (NIC) on the GCSSupercomputer JUWELS at J\"{u}lich Supercomputing Centre (JSC) and through the Center for Information Services and HPC (ZIH) at TU Dresden.

\bibliography{custom}
\bibliographystyle{acl_natbib}

\appendix

\end{document}